\documentclass{article}

\usepackage{PRIMEarxiv}

\usepackage[utf8]{inputenc} 
\usepackage[T1]{fontenc}    
\usepackage[colorlinks,
            linkcolor=blue,       
            anchorcolor=blue,  
            citecolor=blue,    
            ]{hyperref}
\usepackage{url}            
\usepackage{booktabs}       
\usepackage{amsfonts}       
\usepackage{nicefrac}       
\usepackage{microtype}      
\usepackage{lipsum}
\usepackage{fancyhdr}       
\usepackage{graphicx}       
\usepackage{array}       
\graphicspath{{media/}}     
\usepackage{dirtree}
\usepackage{xcolor}
\usepackage{color}
\usepackage{colortbl}
\usepackage{cancel}

\title{KTVIC: A Vietnamese Image Captioning Dataset on the Life Domain 

}

\author{
  Anh-Cuong Pham\thanks{This work was done while Anh-Cuong Pham was a senior student at VNU University of Engineering and Technology.}\\
  VNU University of Engineering and Technology\\
  \texttt{19020038@vnu.edu.vn}
   \And
  Van-Quang Nguyen \\
  RIKEN Center for AIP \\
  \texttt{quang@vision.is.tohoku.ac.jp} \\
    \And
  Thi-Hong Vuong \\
  VNU University of Engineering and Technology \\
  \texttt{hongvt57@vnu.edu.vn} \\
    \And
  Quang-Thuy Ha \\
  VNU University of Engineering and Technology \\
  \texttt{thuyhq@vnu.edu.vn} \\
}

\begin{document}
\maketitle

\begin{abstract}
Image captioning is a crucial task with applications in a wide range of domains, including healthcare and education.
Despite extensive research on English image captioning datasets, the availability of such datasets for Vietnamese remains limited, with only two existing datasets.
In this study, we introduce KTVIC, a comprehensive Vietnamese Image Captioning dataset focused on the life domain, covering a wide range of daily activities. This dataset comprises 4,327 images and 21,635 Vietnamese captions, serving as a valuable resource for advancing image captioning in the Vietnamese language. We conduct experiments using various deep neural networks as the baselines on our dataset, evaluating them using the standard image captioning metrics, including BLEU, METEOR, CIDEr, and ROUGE. Our findings underscore the effectiveness of the proposed dataset and its potential contributions to the field of image captioning in the Vietnamese context.
\end{abstract}


\section{Introduction}

The comprehension of visual context within images constitutes a fundamental objective in the realm of computer vision. This has given rise to image captioning \cite{survey2019, survey2021}, a task dedicated to equipping machines with the capability to decipher contextual information from images and subsequently generate descriptive captions.
The practical applications of image captioning span diverse domains, including its vital role in generating captions for medical images \cite{medicat}. These captions serve as invaluable aids in the process of medical diagnosis and treatment, enabling healthcare professionals to better comprehend and interpret complex imagery, ultimately leading to more accurate diagnoses and improved patient care. Within the educational landscape, this technology serves as a transformative tool by unraveling hidden insights and context within visual content. By providing informative captions for images and illustrations, it empowers students and learners to gain a deeper understanding of the subject matter.


In response to the pressing need for progress in image captioning research, extensive efforts have been devoted to investigating various facets, including data and architectural considerations in the development of automatic image caption generation models \cite{survey2019, survey2021, grit}. Currently, well-established datasets covering diverse subject matter and domains serve as prevalent resources for training image captioning models, exemplified by Visual Genome \cite{visualgenome}. These datasets are also pivotal for the rigorous evaluation of model performance, as demonstrated by the widely-adopted Microsoft COCO dataset \cite{coco}. It is noteworthy that this commitment to fostering image captioning research extends beyond the English language, as evident in the existence of COCO-extended datasets, such as COCO-CN \cite{coco-cn} for Chinese and STAIR \cite{stair} for Japanese, alongside non-COCO-based datasets like PraCegoVer \cite{pracegover} in Portuguese.


However, in stark contrast to this abundance of resources, the Vietnamese language remains inadequately represented in the domain of image captioning, with only two extant datasets: UIT-ViIC \cite{uitviic}, specializing in the sports domain, and VieCap4H \cite{viecap4h}, focusing on the medical domain. Unlike their English counterparts, these Vietnamese datasets operate within a narrower domain, constraining their utility. Consequently, there exists an imperative need to enrich the repository of captioning data, with the aim of improving the quality of image captioning models tailored to the Vietnamese context.

\begin{figure}[!t]
\centerline{\includegraphics[width=\linewidth]{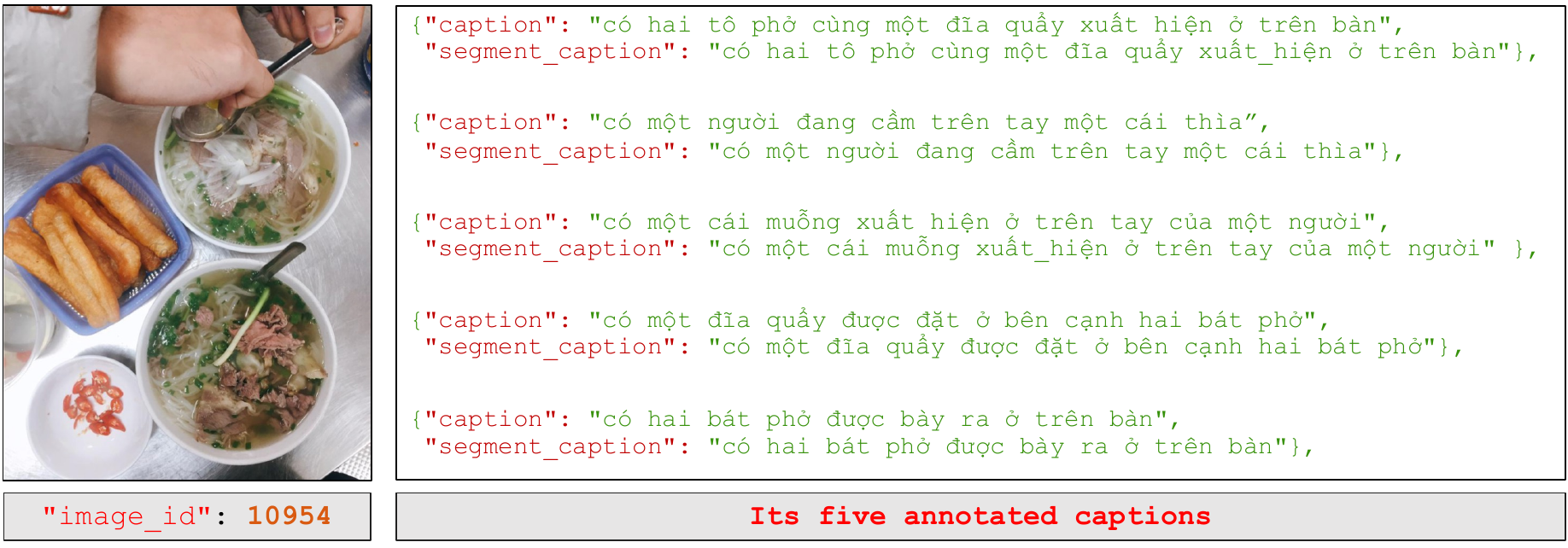}}
\vspace*{8pt}
\caption{An example of image annotation in the KTVIC dataset, where each image is accompanied by five descriptive (segmented) captions.}
\label{fig:example}
\end{figure}

To bridge this gap, we have meticulously created the KTVIC dataset (short for "\textbf{K}nowledge \textbf{T}echnology Lab's \textbf{V}ietnamese \textbf{I}mage \textbf{C}aptioning"). This dataset is centered around the life domain, encompassing various daily activities. 
The images are sourced from the UIT-EVJVQA dataset \cite{vlsp2022}, originally designed for multilingual visual question-answering tasks. We engaged skilled human annotators into the annotation process which follows the established guidelines from prior work \cite{coco, uitviic}. The human annotators were asked to provide each given image with five captions, which describe the visual content in various perspectives; see Figure \ref{fig:example} for an example of data annotation.
This meticulous annotation process resulted in a total of 21,635 high-quality captions annotated by human annotators.

KTVIC distinguishes itself from the two existing image captioning datasets in two crucial aspects. Firstly, the image sources in the life domain are notably more diverse, featuring a richer array of objects within individual images. Secondly, following the standard set by the COCO image captioning dataset \cite{coco}, we provide five captions per image, distinguishing our approach from VieCap4H, in which each image is often annotated with only one caption. We believe that the creation of KTVIC represents a significant contribution that not only addresses the dearth of Vietnamese image captioning resources but also serves as a catalyst for advancing research within this specialized domain.

\section{Related Works}
\label{sec:headings}

Image captioning has seen substantial progress in recent years, driven by the availability of well-established datasets and advancements in model architectures. In this section, we review notable datasets and prior work in the field, both in the context of English datasets and the emerging Vietnamese image captioning datasets.


\subsection{English Datasets}
In the realm of English image captioning research, datasets serve as indispensable tools for advancing the field. Notable among these is the Flickr8k dataset \cite{flickr8kold, flickr8k}, comprising 8,000 images from Flickr, accompanied by approximately 40,000 meticulously crafted captions, thus providing a foundational benchmark for early image captioning studies. An extension of this dataset, Flickr30k \cite{flickr30k}, expands the scope with a larger collection of around 30,000 images and 150,000 associated captions, notable for its diverse scenarios that prove valuable for nuanced image captioning tasks, enabling researchers to explore a broader range of visual contexts. 

The Microsoft COCO dataset \cite{coco}, renowned as a standard benchmark, encompasses over 330,000 images paired with approximately one and a half million captions. Beyond its significance in image captioning, COCO extends its versatility to support various computer vision tasks (e.g., object detection, segmentation) establishing itself as a foundational resource in the landscape of image captioning studies. 

Another prominent dataset, Visual Genome \cite{visualgenome}, plays a crucial role in pre-training object detection to extract region-based features for various vision language models. With over 108,000 images and an extensive collection of over 5 million objects, averaging 40-50 objects per image, Visual Genome provides a rich resource for training models on diverse visual concepts. These datasets collectively contribute to shaping the trajectory of English image captioning research, each offering unique attributes and challenges, thereby fostering a comprehensive understanding of the intricacies within this domain.

\begin{figure}[t]
  \begin{minipage}[b]{0.3\linewidth}
    \centering
    \includegraphics[width=\linewidth]{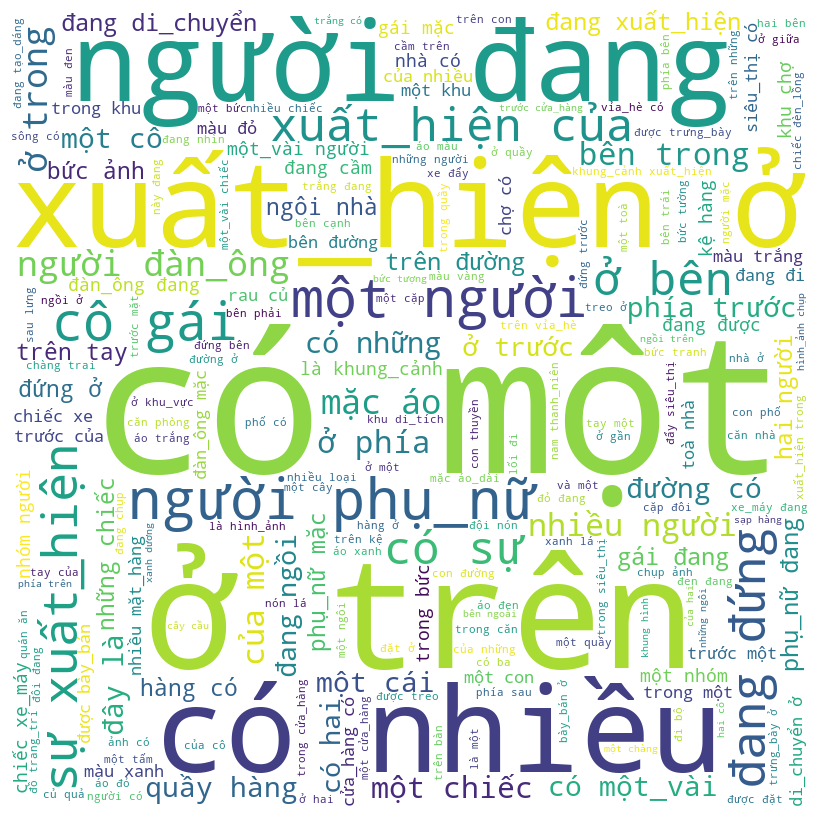}
    \caption{KTVIC word cloud.}
    \label{fig:wordcloud}
  \end{minipage}
  \hspace{0.2cm} 
  \begin{minipage}[b]{0.7\linewidth}
    \centering
    \includegraphics[width=\linewidth]{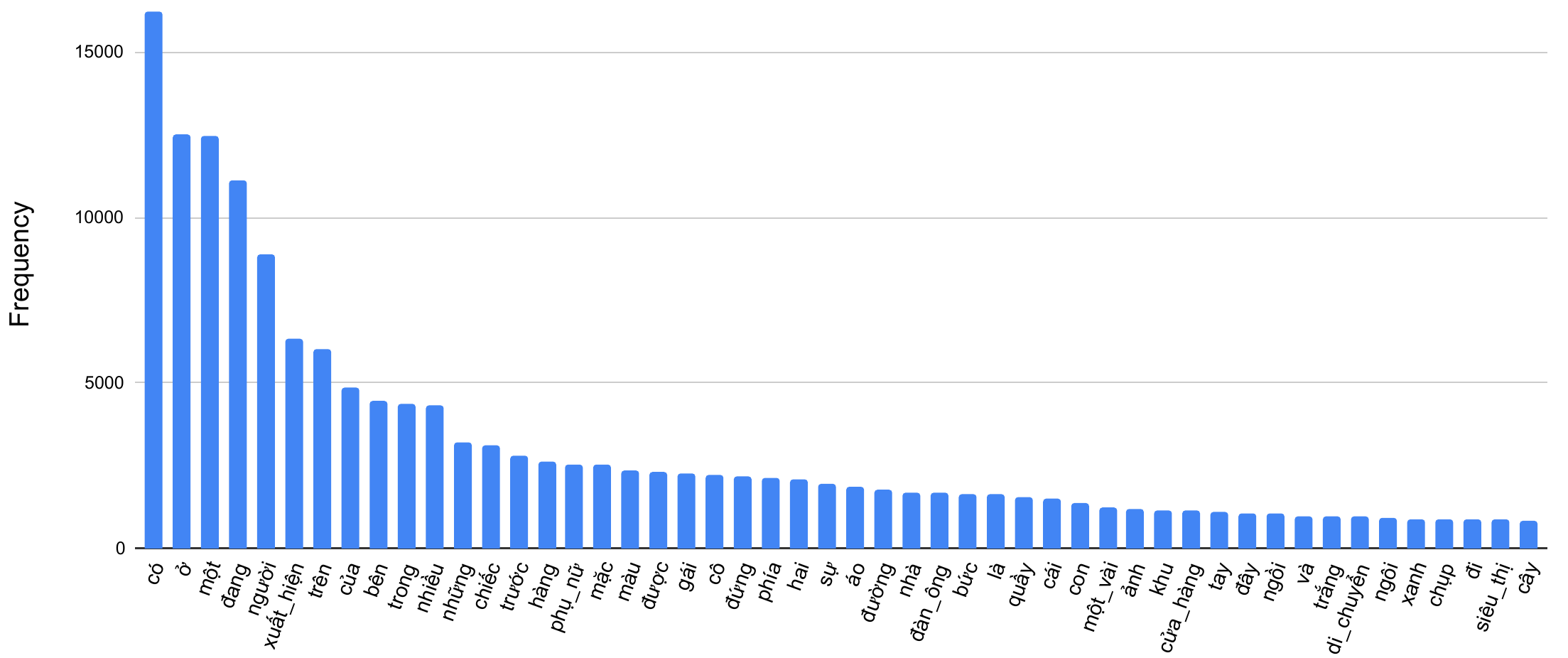}
    \caption{Frequency of top 50 common words in our KTVIC dataset.}
    \label{fig:stat_graph}
  \end{minipage}
\end{figure}

\subsection{Vietnamese Datasets}

In the realm of Vietnamese image captioning research, efforts have been directed towards curating datasets that align with the unique nuances of Vietnamese language. Two distinct approaches have given rise to two noteworthy datasets.

The first approach involves translating existing image captioning datasets from English into Vietnamese and manually evaluating by human. It is exemplified by the UIT-ViIC dataset, the first Vietnamese image captioning dataset, accompanied by the set of caption generation rules proposed by the authors \cite{uitviic}. The UIT-ViIC dataset encompasses 3,850 sports-related images from the COCO dataset, accompanied by 19,250 Vietnamese captions, marking it as the first Vietnamese captioning dataset. While this approach extends the English dataset comprehensively, it is noted that translation tools, though useful in supporting data creation, may not capture the natural flow of Vietnamese due to the nuanced differences between the two languages \cite{uitviic}. Therefore, manual evaluation and filtering remains crucial in ensuring meaningful captions in the target language. 

The second approach involves collecting images from a specific domain and annotate captions for each image manually. An example of this approach is the VieCap4H dataset \cite{viecap4h}, a Vietnamese image captioning dataset focused on the medical domain (including ultrasound, magnetic resonance imaging, optical imaging, etc). VieCap4H was introduced for a contest held by the Vietnamese Language and Speech Processing Association (VLSP). It comprises around 10,000 medical images and 11,000 captions, showcasing its relevance in the medical field, particularly in the context of and post the pandemic. However, mitigating potential biases in the medical field requires careful consideration that involves expert knowledge in the medical domain.

In this study, we contribute to Vietnamese image captioning by introducing the KTVIC dataset, which focuses on the life domain. Our dataset creation aligns with the second approach, yet with two significant differences. Firstly, the image sources in the life domain are notably more diverse, featuring a richer array of objects within individual images. Specifically, we leverage images from the UIT-EVJVQA dataset \cite{vlsp2022}. This dataset, centered around life-related content, comprises approximately 5,000 images with 30,000 question-answer pairs in three languages: Vietnamese, English, and Japanese. It was originally designed for the MVQA task in the VLSP 2022 challenge. We opted for these image sources for caption annotation due to their inclusion of multiple objects in each image, a departure from the characteristics of the two previous captioning datasets (UIT-ViIC and VieCap4H), which often exhibit only a limited number of objects per image. Secondly, following to the standard set by the COCO image captioning dataset \cite{coco}, we provide five captions per image, distinguishing our approach from VieCap4H, in which each image is often annotated with only one caption.

\section{KTVIC: Vietnamese Image Captioning Dataset}

\label{sec:others}





\subsection{Specifications}
\begin{table}[t]
    \centering
    \begin{minipage}{.45\linewidth}
        \centering
        \setlength{\extrarowheight}{3pt}
        \caption{The splits in KTVIC}
        \begin{tabular}{l|cc}
            \textbf{Split} & \textbf{Images} & \textbf{Captions} \\
            \hline
            Train & 3,769 & 18,845 \\
            Test & 558 & 2,790 \\
        \end{tabular}
        \label{tab:split}
    \end{minipage}%
    \begin{minipage}{.45\linewidth}
        \centering
        \setlength{\extrarowheight}{3pt}
        \caption{Token and word counts in KTVIC}
        \begin{tabular}{l|cc}
            \textbf{Type} & \textbf{Total} & \textbf{Unique} \\
            \hline
            Tokens & 274,204 & 1,796 \\
            Words & 237,429 & 2,400 \\
        \end{tabular}
        \label{tab:token_word}
    \end{minipage}
\end{table}

The KTVIC dataset offers caption annotations for 4,327 images publicly sourced from the UIT-EVJVQA dataset, resulting in a total of 21,635 captions. Each image, denoted as $x$, is annotated with \textbf{five captions} by trained human annotators. Notably, having five captions per image allows for the capture of visual content from different perspectives, contributing to the linguistic diversity of the dataset. These annotations encapsulate a diverse array of scenes depicting the daily activities of Vietnamese people or various locations within Vietnam \cite{vlsp2022}.

Adhering to the initial splits in the UIT-EVJVQA dataset, the 4,327 images are divided into train and test sets, comprising 3,769 and 558 images, respectively; see Table \ref{tab:split} for more details. This division preserves the original splits, maintaining consistency with the dataset's structure. The dataset annotation is provided at our GitHub website \footnote{\url{https://github.com/pacman-ctm/ktvic}}. The annotation of the test and training splits is provided with two separate JSON files, which are structured in the format borrowed from the COCO image captioning dataset. Given the nature of the Vietnamese language, where a word can consist of one or multiple tokens, we provide the segmented captions (denoted as $\texttt{"segment\_caption"}$) alongside the annotated captions (refer to Figure \ref{fig:example}). 
To obtain the segmented captions, we utilize RDRSegmenter, a highly accurate Vietnamese word segmenter \cite{rdrsegmenter}. 

Table \ref{tab:token_word} provides insights into the linguistic distribution within the annotated (segmented) captions of the dataset. Specifically, the dataset comprises a total of 274,204 tokens across all captions, featuring by 1,796 unique tokens. The segmented version consists of 237,429 words from 2,400 unique words. This underscores the lexical richness and variety inherent in the language used to annotate descriptive captions for the images. 
Figure \ref{fig:caption_length} illustrates the distribution of caption length based on the number of words in a segmented caption. The captions exhibit lengths ranging from 5 to 27 words, with prominent peaks observed between 8 and 13 words. On average, the length of segmented captions is 10.97 words.

{\color{blue} 

\begin{figure}[t]
\centerline{\includegraphics[width=0.8\linewidth]{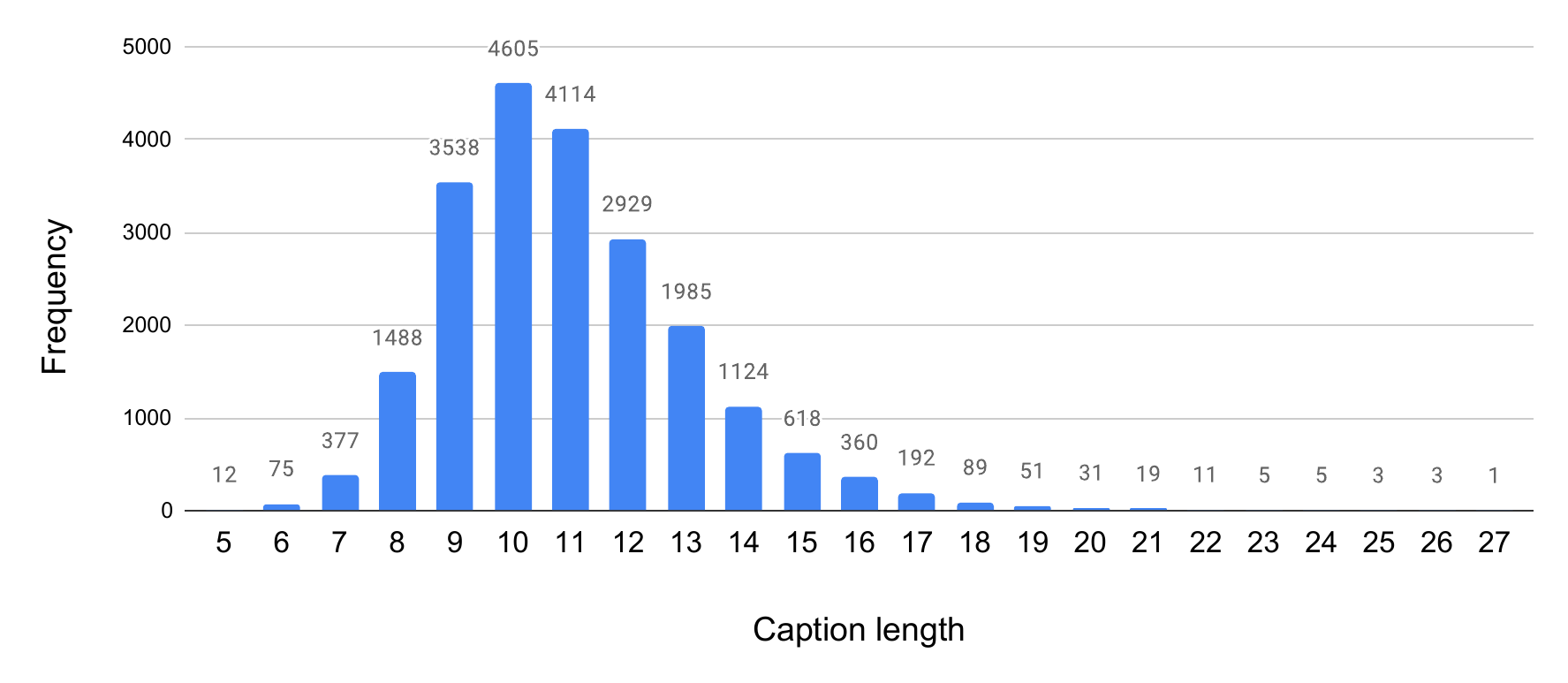}}

\vspace*{8pt}
\caption{Caption length in terms of the number of words in captions}
\label{fig:caption_length}
\end{figure}
}

\subsection{Annotation Process}
\begin{figure}[!th]
\centerline{\includegraphics[width=0.85\linewidth]{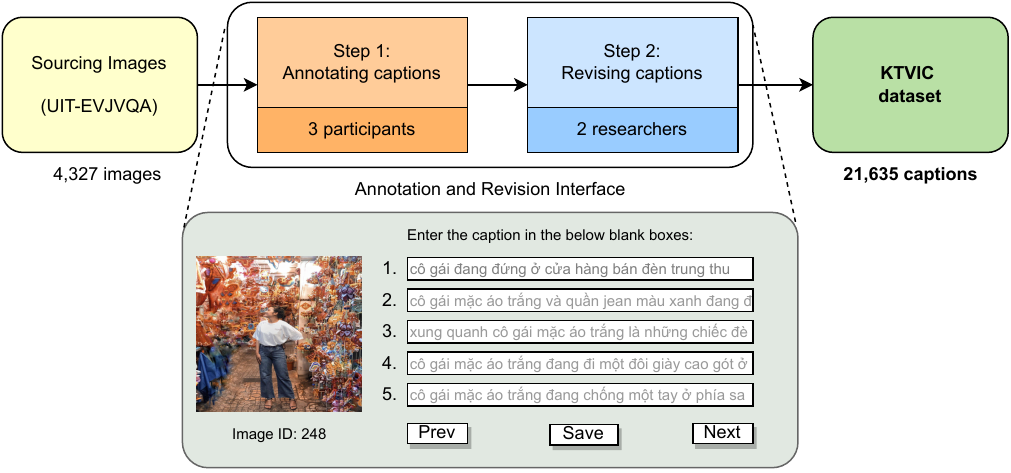}}
\vspace*{8pt}
\caption{The annotation process and the simple interface for human annotators to caption images in the annotation and revising steps.}
\label{fig:process}
\end{figure}

We construct the KTVIC dataset following the annotation process and established guidelines from previous work \cite{coco, uitviic}. Figure \ref{fig:process} depicts the annotation process which unfolds in two phases: caption annotation and caption revision. 
Before starting the annotation process, we provided information and training to the human annotators participating in the task, familiarizing them with the annotation rules.

Originally, UIT-ViIC defined ten annotation rules for generating captions, drawing inspiration from the MS-COCO annotation rules \cite{coco, uitviic}. These ten rules are outlined below:
\begin{enumerate}
    \item Ensure each caption consists of a minimum of ten Vietnamese words.
    \item Only describe visible activities and objects included in the image.
    \item Exclude names of places, streets (e.g., Chinatown, New York), and numerical details (e.g., apartment numbers, specific TV times).
    \item Allow the use of familiar English words like laptop, TV, tennis, etc.
    \item Structure each caption as a single sentence in continuous tense.
    \item Omit personal opinions and emotions from annotations.
    \item Permit annotators to describe activities and objects from various perspectives.
    \item Focus solely on describing visible "thing" objects.
    \item Disregard ambiguous "stuff" objects lacking clear borders
    \item If there are 10 to 15 objects of the same category or species, annotators may omit them in captions.
\end{enumerate}

However, after analyzing UIT-ViIC and VieCap4H datasets, we found that Rule 1 imposes constraints on the flexibility of manual caption annotation for images, limiting the diversity of generated sentences. Consequently, we decided to \textbf{disregard Rule 1 while upholding the remaining nine rules for our annotation process}. This is believed to facilitate the annotation process of accurate and diverse captions for images in the dataset. Additionally, we leverage our general knowledge and acquire essential information and attributes of annotated images in advance by examining question-answer pairs from the UIT-EVJVQA dataset. Then, we designed a minimal interface shown in Figure \ref{fig:process} for captioning images, following the approach outlined in \cite{uitviic}, aiming at minimizing distractions.

In the initial phase, three participants follow the nine rules to annotate captions for the images sourced from the UIT-EVJVQA dataset. The annotation interface presents human annotator an image from the UIT-EVJVQA dataset to provide five captions in the corresponding blank boxes. The second phase engages two experienced researchers who review and revise five captions given its image. The researchers rectified any linguistic errors or those that violate the annotation rules. 

In total, we annotated 4,327 images (comprising 3,769 images in the training set and 558 images in the test set), with each image having 5 high-quality captions. This resulted in a total of 21,635 captions annotated by human annotators.

\begin{figure}[t]
\centerline{\includegraphics[width=\linewidth]{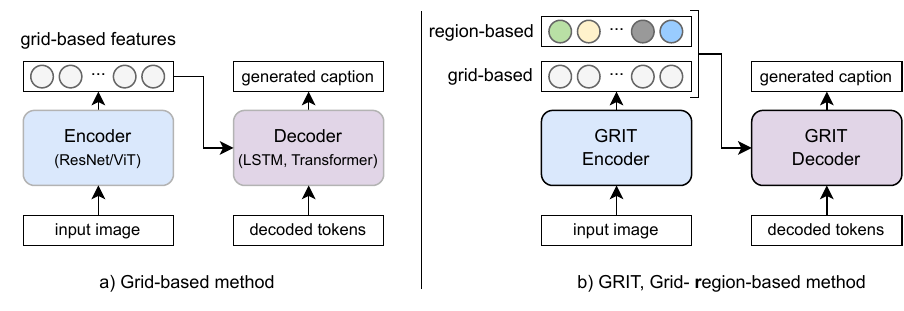}}
\vspace*{8pt}
\caption{Baseline methods experimented on the KTVIC dataset: a) Grid-based methods utilizing grid-based features from ResNet/ViT to generate captions; b) GRIT, an advanced method that leverages both grid-based and region-based features for caption generation.}
\label{fig:methods}
\end{figure}

\section{Baseline Methods on KTVIC} \label{sec3}


Recent image captioning methods commonly adopt an encoder-decoder architecture. In this framework, the encoder processes an image to extract visual features, and the decoder, utilizing these features, generates a sequence of words in an auto-regressive manner, thereby forming a descriptive caption for the image.

The key to superior performance lies in addressing the challenge of extracting good visual features. Based on the feature extraction, we can categorize the image captioning methods into three categories: i) grid-based methods which utilize local features extracted at the regular grid points, often obtained directly from a higher layer feature map(s) of CNNs/ViTs \cite{vinyals2015show,lu2017knowing} ii) or region-based methods \cite{anderson2018bottom} which utilizes set of local image features of the regions detected from pretrained object detectors. Grid features are local image features extracted at the regular grid points,
often obtained directly from a higher layer feature map(s) of CNNs/ViTs. Still, region-based features have become the de-facto choice of visual representation for image captioning although coming with the high computational cost.

In this research, we explore two grid-based methods (denoted as Baseline 1 and Baseline 2) for their computational efficiency, considering our limited budget. Furthermore, we conduct experiments using GRIT \cite{grit} as Baseline 3, an advanced image captioning method that has demonstrated state-of-the-art performance on benchmarks for English-based image captioning. In total, we establish three baselines on our proposed dataset as shown on Figure \ref{fig:methods}, and detailed information is provided below.

\subsection{Baseline 1 with a CNN Encoder and LSTM Decoder}
We establish the initial baseline employing a grid-based architecture with ResNet \cite{residual} as the feature extractor. ResNet is a deep residual convolutional neural network \cite{cnn} constructed with a series of residual blocks, incorporating convolutional layers and a skip connection. The feature maps extracted from the final layer of ResNet are then input into the LSTM decoder \cite{lstm} to generate a sequence of words in an auto-regressive manner.


\subsection{Baseline 2 with a ViT Encoder and Transformer Decoder}
The second baseline also utilizes a grid-based architecture, employing the Vision Transformer (ViT) as the feature extractor. ViTs \cite{attention} have demonstrated superior performance in image recognition and are increasingly applied to various problems in Computer Vision. In this baseline, we leverage the grid-based features extracted from the final layer of ViT, passing them into a more advanced decoder composed of several transformer layers \cite{attention}.  The decision to use the Transformer decoder stems from its efficiency and accuracy in caption generation, as highlighted in the literature \cite{survey2021}. The decoder operates in an auto-regressive manner, generating words to form the best caption based on the features extracted from ViT.

\subsection{Baseline 3 using GRIT Architecture}

The third baseline explored in our experiments on the proposed dataset is GRIT \cite{grit}, a method that fuses grid- and region-based features for image captioning. GRIT has not only established a new state-of-the-art standard in image captioning benchmarks such as COCO but also addresses computational challenges posed by conventional region-based methods. It achieves this by incorporating a DETR-like detector into its design, creating a monolithic architecture.
Specifically, the model leverages Swin Transformer \cite{swin} (an advanced variant of ViTs) as the visual backbone to extract features from the input image. Subsequently, these visual features are directed to two distinct networks based on the Transformer encoder—one focused on extracting region features and the other on extracting grid features. Both sets of visual features are then channeled to the caption generator, based on the Transformer decoder using parallel attention mechanism \cite{attn}.


\section{Experiments on KTVIC}

The specific implementation details for the three baselines are outlined below.
\begin{itemize}
    \item Baseline 1: We utilize a ResNet101 model pre-trained on ImageNet \cite{deng2009imagenet} as the feature extractor, coupled with a decoder comprising a single LSTM layer. The entire model is fine-tuned over 10 epochs using the Adam optimizer \cite{adam}, with a learning rate set to 10e-3.
    
    \item Baseline 2: We employ a Vision Transformer (ViT-base) model pre-trained on ImageNet \cite{deng2009imagenet} as the feature extractor. The decoder consists of 1 transformer layer. The entire model undergoes fine-tuning for 10 epochs using the Adam optimizer with a learning rate set to 10e-4.
    
    \item Baseline 3: We directly employ the GRIT architecture, utilizing their pre-trained backbone and object detector network. The model is fine-tuned on our dataset for 10 epochs, employing the Adam optimizer with a learning rate set to 5 $\times$ 10e-6. Additionally, we implement beam search with a beam size of 5 for caption generation.
\end{itemize}

It is important to highlight that, for simplicity and reproducibility, all the baselines are fine-tuned using a cross-entropy loss (XE) without additional refinement through CIDEr-D optimization with self-critical sequence training strategy \cite{cideropt}. The source code for the three baseline models is available in our GitHub repository \footnote{Our code is accessible at: \url{https://github.com/pacman-ctm/thesis-code} (for the first two baselines) and \url{https://github.com/davidnvq/grit/tree/vicap} (for the GRIT-based baseline).}.


\subsection{Evaluation Metrics}
We employ four standard metrics widely used in image captioning: BLEU-1 and BLEU-4 \cite{bleu}, METEOR \cite{meteor}, ROUGE-L \cite{rouge}, and CIDEr \cite{cider}. These metrics evaluate the similarity between the generated description and its ground truth, the segmented caption.
Specifically, BLEU measures the similarity based on n-gram matches between reference sentences and generated sentences. METEOR is calculated by comparing and combining information from both the translated and original text. The ROUGE-L score measures the overlap between the longest common subsequence of two sentences. On the other hand, CIDEr evaluates the similarity based on overlapping natural language words, comparing human-generated captions with model-generated captions. Conventionally, CIDEr is often considered as the most important metric for comparison in image captioning.


\subsection{Experimental Results}


\begin{table}[t]
    \setlength{\extrarowheight}{3pt}
    \caption{Quantitative results on the test split of KTVIC. Higher values indicate better for all metrics.}
    \centering
    \begin{tabular}{lcc|cccccc}
        \textbf{Model} & \textbf{Encoder} & \textbf{Decoder} & \textbf{BLEU-1} & \textbf{BLEU-4} & \textbf{METEOR} & \textbf{ROUGE} & \textbf{CIDEr}  \\ \hline
        1) & ResNet101 & LSTM & 53.0 & 15.5 & 27.5 & 42.3 & 21.8 \\
        2) & ViT & Transformer & 56.1 & 19.3 & 30.1 & 44.7 & 36.8\\
        3) & GRIT & GRIT  & \textbf{74.7} & \textbf{40.6} & \textbf{36.6} & \textbf{59.7} & \textbf{136.0} \\
    \end{tabular}
    \label{table:quantitative}
\end{table}

Table \ref{table:quantitative} presents the performance of the three baselines on the test split of KTVIC. All three models demonstrate the capability to generate positive evaluation metrics for Vietnamese captions, underscoring KTVIC as a valuable and effective dataset for Vietnamese Image Captioning.

Notably, the results from the two Transformer-based models surpass those of the model without the Transformer architecture (Baseline 1 that uses ResNet101 and LSTM). Furthermore, the GRIT-based model outperforms the other baselines with a large margin in all evaluation metrics, especially CIDEr and BLEU scores. This suggests Transformer-based models deliver favorable results in Vietnamese Image Captioning tasks, and combining grid and region features proves more effective than relying solely on region features.

\begin{figure}[ht!]
    \centering
        \includegraphics[width=0.8\linewidth]{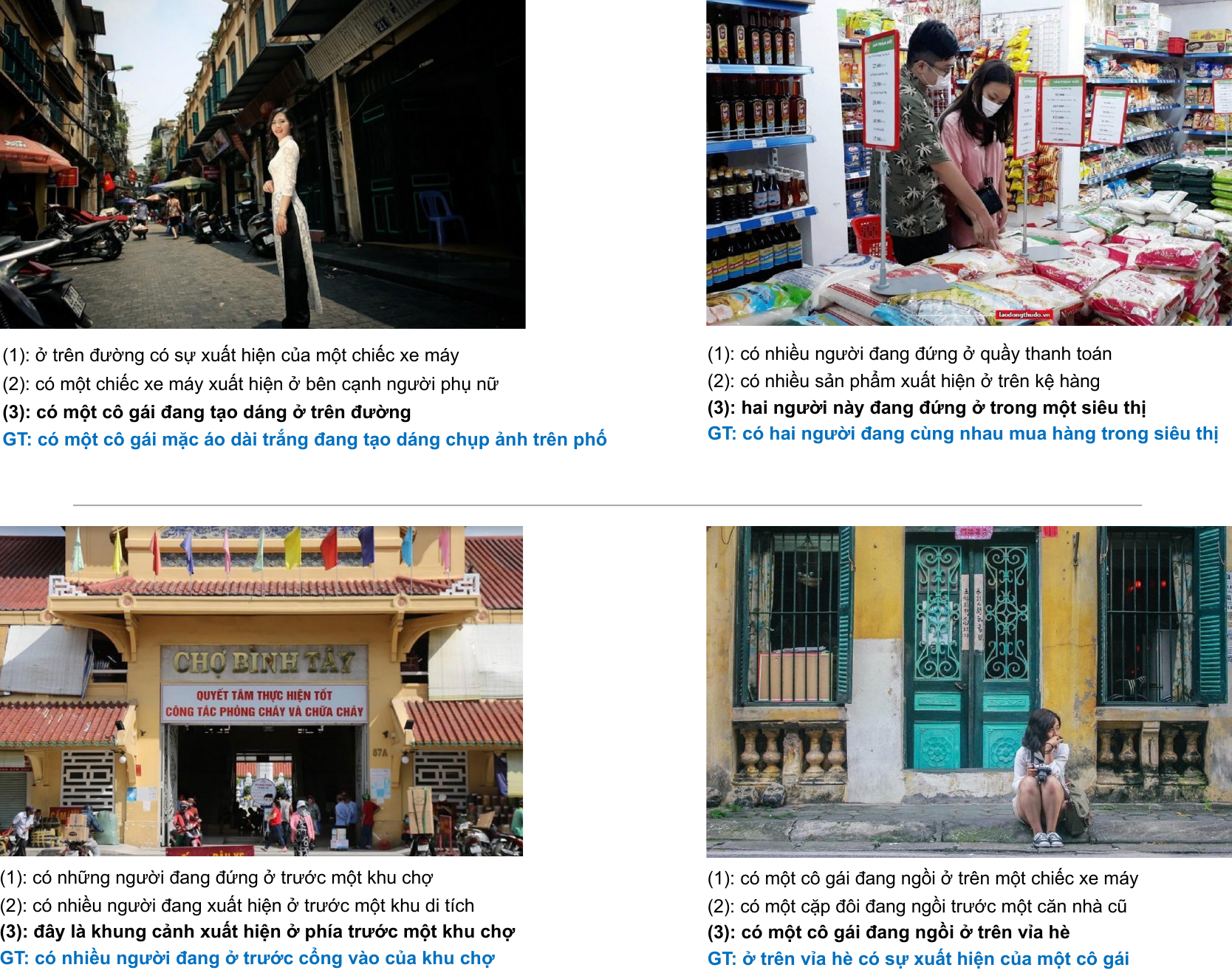}
    \caption{Examples of inference images from the KTVIC dataset with our baselines: Baseline (1) using ResNet \& LSTM), Baseline (2) using ViT \& Transformer decoder, and Baseline (3) -- GRIT, alongside the ground truth captions for each image (Zoom in for a better view).}
    \label{fig:qualitative}
\end{figure}

Figure \ref{fig:qualitative} depicts the qualitative results achieved by the three baselines on the test images. It is seen that Transformer-based methods (Baseline 2 and Baseline 3) alleviate the errors of misidentifying objects observed in Baseline 1. Notably, Baseline 3 (GRIT) exhibits the most promising qualitative results, producing high-quality semantic captions close to the ground truth descriptions provided by human annotators. Despite these positive results, we believe that further research is essential to advance the image captioning task specifically tailored for the Vietnamese context.

\section{Conclusions}



In this study, we have introduced KTVIC, a novel Vietnamese image captioning dataset that focuses on the life domain. Comprising 4,327 images and 21,635 Vietnamese captions, this dataset was carefully curated by human annotators to enhance resources for image captioning tasks in the Vietnamese language. KTVIC distinguishes itself from existing datasets in two crucial aspects: it covers a diverse range of daily activities, featuring a richer array of objects within individual images, and it provides five captions per image, offering more semantic signals to enhance model training.

Furthermore, we experimented with several baselines employing deep neural network architectures and conducted evaluations on the test split using standard image captioning metrics. The quantitative and qualitative results obtained on the test images highlight the potential utility of our proposed dataset as a valuable resource for advancing image captioning in the Vietnamese language.



\section*{Acknowledgments}
We extend our gratitude to the dedicated efforts of lecturers and students at the Data Science and Knowledge Technology Laboratory, VNU University of Engineering and Technology, for their valuable support in the creation and testing of the KTVIC dataset.






\bibliographystyle{unsrt}  
\bibliography{references}

\begin{thebibliography}{10}

\bibitem{survey2019}
Shuang Liu, Liang Bai, Yanli Hu, and Haoran Wang.
\newblock Image captioning based on deep neural networks.
\newblock In {\em MATEC web of conferences}, volume 232, page 01052. EDP Sciences, 2018.

\bibitem{survey2021}
Matteo Stefanini, Marcella Cornia, Lorenzo Baraldi, Silvia Cascianelli, Giuseppe Fiameni, and Rita Cucchiara.
\newblock From show to tell: A survey on deep learning-based image captioning.
\newblock {\em IEEE transactions on pattern analysis and machine intelligence}, 45(1):539--559, 2022.

\bibitem{medicat}
Sanjay Subramanian, Lucy~Lu Wang, Ben Bogin, Sachin Mehta, Madeleine van Zuylen, Sravanthi Parasa, Sameer Singh, Matt Gardner, and Hannaneh Hajishirzi.
\newblock Medicat: A dataset of medical images, captions, and textual references.
\newblock In {\em Findings of the Association for Computational Linguistics: EMNLP 2020}, pages 2112--2120, 2020.

\bibitem{grit}
Van-Quang Nguyen, Masanori Suganuma, and Takayuki Okatani.
\newblock Grit: Faster and better image captioning transformer using dual visual features.
\newblock In {\em European Conference on Computer Vision}, pages 167--184. Springer, 2022.

\bibitem{visualgenome}
Ranjay Krishna, Yuke Zhu, Oliver Groth, Justin Johnson, Kenji Hata, Joshua Kravitz, Stephanie Chen, Yannis Kalantidis, Li-Jia Li, David~A Shamma, et~al.
\newblock Visual genome: Connecting language and vision using crowdsourced dense image annotations.
\newblock {\em International journal of computer vision}, 123:32--73, 2017.

\bibitem{coco}
Tsung-Yi Lin, Michael Maire, Serge Belongie, James Hays, Pietro Perona, Deva Ramanan, Piotr Doll{\'a}r, and C~Lawrence Zitnick.
\newblock Microsoft coco: Common objects in context.
\newblock In {\em Computer Vision--ECCV 2014: 13th European Conference, Zurich, Switzerland, September 6-12, 2014, Proceedings, Part V 13}, pages 740--755. Springer, 2014.

\bibitem{coco-cn}
Xirong Li, Chaoxi Xu, Xiaoxu Wang, Weiyu Lan, Zhengxiong Jia, Gang Yang, and Jieping Xu.
\newblock Coco-cn for cross-lingual image tagging, captioning, and retrieval.
\newblock {\em IEEE Transactions on Multimedia}, 21(9):2347--2360, 2019.

\bibitem{stair}
Yuya Yoshikawa, Yutaro Shigeto, and Akikazu Takeuchi.
\newblock Stair captions: Constructing a large-scale japanese image caption dataset.
\newblock {\em arXiv preprint arXiv:1705.00823}, 2017.

\bibitem{pracegover}
Gabriel~Oliveira dos Santos, Esther~Luna Colombini, and Sandra Avila.
\newblock \# pracegover: A large dataset for image captioning in portuguese.
\newblock {\em Data}, 7(2):13, 2022.

\bibitem{uitviic}
Quan~Hoang Lam, Quang~Duy Le, Van~Kiet Nguyen, and Ngan Luu-Thuy Nguyen.
\newblock Uit-viic: A dataset for the first evaluation on vietnamese image captioning.
\newblock In {\em International Conference on Computational Collective Intelligence}, pages 730--742. Springer, 2020.

\bibitem{viecap4h}
Dang~Long Hoang, Nguyen~Thanh Son, Nguyen Thi~Minh Huyen, Vu~Xuan Son, et~al.
\newblock Vlsp 2021-viecap4h challenge: Automatic image caption generation for healthcare domain in vietnamese.
\newblock {\em VNU Journal of Science: Computer Science and Communication Engineering}, 38(2), 2022.

\bibitem{vlsp2022}
Ngan Luu-Thuy Nguyen, Nghia~Hieu Nguyen, Duong~TD Vo, Khanh~Quoc Tran, and Kiet Van~Nguyen.
\newblock Vlsp 2022--evjvqa challenge: Multilingual visual question answering.
\newblock {\em arXiv preprint arXiv:2302.11752}, 2023.

\bibitem{flickr8kold}
Micah Hodosh, Peter Young, and Julia Hockenmaier.
\newblock Framing image description as a ranking task: Data, models and evaluation metrics.
\newblock {\em Journal of Artificial Intelligence Research}, 47:853--899, 2013.

\bibitem{flickr8k}
Micah Hodosh, Peter Young, and Julia Hockenmaier.
\newblock Framing image description as a ranking task: Data, models and evaluation metrics (extended abstract).
\newblock In Qiang Yang and Michael~J. Wooldridge, editors, {\em Proceedings of the Twenty-Fourth International Joint Conference on Artificial Intelligence, {IJCAI} 2015, Buenos Aires, Argentina, July 25-31, 2015}, pages 4188--4192. {AAAI} Press, 2015.

\bibitem{flickr30k}
Bryan~A Plummer, Liwei Wang, Chris~M Cervantes, Juan~C Caicedo, Julia Hockenmaier, and Svetlana Lazebnik.
\newblock Flickr30k entities: Collecting region-to-phrase correspondences for richer image-to-sentence models.
\newblock In {\em Proceedings of the IEEE international conference on computer vision}, pages 2641--2649, 2015.

\bibitem{rdrsegmenter}
Dat~Quoc Nguyen, Thanh Vu, Mark Dras, Mark Johnson, et~al.
\newblock A fast and accurate vietnamese word segmenter.
\newblock In {\em Proceedings of the Eleventh International Conference on Language Resources and Evaluation (LREC 2018)}, 2018.

\bibitem{vinyals2015show}
Oriol Vinyals, Alexander Toshev, Samy Bengio, and Dumitru Erhan.
\newblock Show and tell: A neural image caption generator.
\newblock In {\em Proceedings of the IEEE Conference on Computer Vision and Pattern Recognition}, pages 3156--3164, 2015.

\bibitem{lu2017knowing}
Jiasen Lu, Caiming Xiong, Devi Parikh, and Richard Socher.
\newblock Knowing when to look: Adaptive attention via a visual sentinel for image captioning.
\newblock In {\em Proceedings of the IEEE Conference on Computer Vision and Pattern Recognition}, pages 375--383, 2017.

\bibitem{anderson2018bottom}
Peter Anderson, Xiaodong He, Chris Buehler, Damien Teney, Mark Johnson, Stephen Gould, and Lei Zhang.
\newblock Bottom-up and top-down attention for image captioning and visual question answering.
\newblock In {\em Proceedings of the IEEE conference on computer vision and pattern recognition}, pages 6077--6086, 2018.

\bibitem{residual}
Kaiming He, Xiangyu Zhang, Shaoqing Ren, and Jian Sun.
\newblock Deep residual learning for image recognition.
\newblock In {\em Proceedings of the IEEE conference on computer vision and pattern recognition}, pages 770--778, 2016.

\bibitem{cnn}
Kunihiko Fukushima.
\newblock Neocognitron: A self-organizing neural network model for a mechanism of pattern recognition unaffected by shift in position.
\newblock {\em Biological cybernetics}, 36(4):193--202, 1980.

\bibitem{lstm}
Sepp Hochreiter and J{\"u}rgen Schmidhuber.
\newblock Long short-term memory.
\newblock {\em Neural computation}, 9(8):1735--1780, 1997.

\bibitem{attention}
Ashish Vaswani, Noam Shazeer, Niki Parmar, Jakob Uszkoreit, Llion Jones, Aidan~N Gomez, {\L}ukasz Kaiser, and Illia Polosukhin.
\newblock Attention is all you need.
\newblock {\em Advances in neural information processing systems}, 30, 2017.

\bibitem{swin}
Ze~Liu, Yutong Lin, Yue Cao, Han Hu, Yixuan Wei, Zheng Zhang, Stephen Lin, and Baining Guo.
\newblock Swin transformer: Hierarchical vision transformer using shifted windows.
\newblock In {\em Proceedings of the IEEE/CVF international conference on computer vision}, pages 10012--10022, 2021.

\bibitem{attn}
Van-Quang Nguyen, Masanori Suganuma, and Takayuki Okatani.
\newblock Efficient attention mechanism for visual dialog that can handle all the interactions between multiple inputs.
\newblock In {\em Computer Vision--ECCV 2020: 16th European Conference, Glasgow, UK, August 23--28, 2020, Proceedings, Part XXIV 16}, pages 223--240. Springer, 2020.

\bibitem{deng2009imagenet}
Jia Deng, Wei Dong, Richard Socher, Li-Jia Li, Kai Li, and Li~Fei-Fei.
\newblock Imagenet: A large-scale hierarchical image database.
\newblock In {\em 2009 IEEE conference on computer vision and pattern recognition}, pages 248--255. Ieee, 2009.

\bibitem{adam}
Diederik~P Kingma and Jimmy Ba.
\newblock Adam: A method for stochastic optimization.
\newblock {\em arXiv preprint arXiv:1412.6980}, 2014.

\bibitem{cideropt}
Steven~J Rennie, Etienne Marcheret, Youssef Mroueh, Jerret Ross, and Vaibhava Goel.
\newblock Self-critical sequence training for image captioning.
\newblock In {\em Proceedings of the IEEE conference on computer vision and pattern recognition}, pages 7008--7024, 2017.

\bibitem{bleu}
Kishore Papineni, Salim Roukos, Todd Ward, and Wei-Jing Zhu.
\newblock Bleu: a method for automatic evaluation of machine translation.
\newblock In {\em Proceedings of the 40th annual meeting of the Association for Computational Linguistics}, pages 311--318, 2002.

\bibitem{meteor}
Satanjeev Banerjee and Alon Lavie.
\newblock Meteor: An automatic metric for mt evaluation with improved correlation with human judgments.
\newblock In {\em Proceedings of the acl workshop on intrinsic and extrinsic evaluation measures for machine translation and/or summarization}, pages 65--72, 2005.

\bibitem{rouge}
Chin-Yew Lin.
\newblock {ROUGE}: A package for automatic evaluation of summaries.
\newblock In {\em Text Summarization Branches Out}, pages 74--81, Barcelona, Spain, July 2004. Association for Computational Linguistics.

\bibitem{cider}
Ramakrishna Vedantam, C.~Lawrence Zitnick, and Devi Parikh.
\newblock Cider: Consensus-based image description evaluation.
\newblock In {\em 2015 IEEE Conference on Computer Vision and Pattern Recognition (CVPR)}, pages 4566--4575, 2015.

\end{thebibliography}

\end{document}